\documentstyle[aclap]{article}

\input psfig.sty

\author{Jo Calder \\
University of Edinburgh\\ 
Language Technology Group, Human Communication Research Centre and\\
Centre for Cognitive Science \\
2 Buccleuch Place \\
Edinburgh, Scotland EH8 9LW \\
\texttt{J.Calder@ed.ac.uk}
}

\begin{document}

\title{On aligning trees}

\maketitle

\begin{abstract}
  The increasing availability of corpora annotated for linguistic
  structure prompts the question: if we have the same texts, annotated
  for phrase structure under two different schemes, to what extent do
  the annotations agree on structuring within the text?  We suggest
  the term \emph{tree alignment} to indicate the situation where two
  markup schemes choose to bracket off the same text elements.  
  We propose a general method for determining agreement between two
  analyses.  We then describe an efficient implementation, which is
  also modular in that the core of the implementation can be reused
  regardless of the format of markup used in the corpora.  The output
  of the implementation on the Susanne and Penn treebank corpora is
  discussed. 
\end{abstract}

\section{Introduction}

We present here a general design for, and modular implementation of,
an algorithm for computing areas of agreement between structurally
annotated corpora.
  Roughly speaking, if two corpora  bracket off the
same stretches of words in their structural analysis of a text, the
corpora agree that that stretch of text should be considered a single
unit at some level of structure.  We will (borrowing a usage from
\cite{Church+Gale:AS} term this agreement (sub)tree
alignment.  

We make the following assumptions, which appear reasonable for markup
schemes with which we are familiar:
\begin{itemize}
\item the ``content'' of each text consists of a sequence of ``terminal''
elements.   That is, the content is a collection of elements generally
corresponding to words and punctuation and this will be roughly
constant across the two corpora.  It may also contain additional
elements to represent, for example, the positing of orthographically
null categories.  

\item the two corpora whose trees are to be aligned contain identifiable
structural markup.  That is, structural ``delimiters'' are distinct
from other forms of markup and content.

\item two corpora agree on an analysis when they bracket off the same
  content. 

\item The corpora may contain additional markup provided this is distinct
from content and structural markup.  
\end{itemize}

Our goal, then, is to determine those stretches of a text's content
which two corpora agree on.  Why might we want to do this?  There are
several reasons:

\begin{itemize}
\item increase confidence in markup and determine areas of
  disagreement 

If two or more corpora agree on parts of an analysis, one may
``trust'' that choice of grouping more than those groupings on which
the corpora differ.   Alignment can be used to detect disagreements
between manual annotators. 

\item verify preservation of analyses across multiple versions of a corpus 

If all the subtrees of a corpus are aligned with those of another, then
the second is consistent with the first, and represents analyses at least as
detailed as those in the first.  Such automatic checking will be
useful both in the case of manual edits to a corpus, and also in the
case where automatic analysis is performed.  

\item import markup from one corpus to another

If one corpus contains ``richer'' information than another, for
example in terms of annotation of syntactic function or of lexical
category, the markup from the first may be interpreted with respect to
analyses in the second. 

\item determine constant markup transformations

Having identified aligned subtrees, the labels of a pair of trees may
be recorded, and the results for the pair of corpora analysed to
determine consistent differences in markup.  

\item determine constant tree transformations

A set of pairings between aligned subtrees can be used as a bootstrap
for semi-automatic markup of corpora.

\end{itemize}

We can also identify some specific motivations
and applications.  First, in the automatic determination of
subcategorization information, confidence in the choice of
subcategorization may be improved by analyses which confirm that
subcategorization from other corpora.  Second, the algorithm we have
developed is robust in the face of minor editorial differences, choice
of markup for punctuation, and overall presentation of the corpora.
We have processed the Susanne corpus \cite{sampson:susanne} and Penn treebank
\cite{PennTB}
to provide
tables of word and subtree alignments.  Third, on the basis of the
computed alignments between the two corpora, and the tree
transformations they imply, the possibility is now open to produce,
semi-automatically, versions of those parts of the Brown corpus
covered by the Penn treebank but not by Susanne, in a Susanne-like
format. Finally, in the development of phrasal parsers, our results
can be used to obtain a measure of how contentious the analysis of
different phrase types is.

Obviously, the utility of algorithms such as the one we present here
is dependent on the quality and reliability of markup in the corpora we
process.

\section{The Task}

In this section, we provide a general characterization of agreement in
analysis between two corpora.  

We assume the existence of two corpora, $C^l$ and $C^r$\footnote{for
  \emph{left} and \textit{right}.}.  The contents of each corpus is a
  sequence of elements drawn from a collection of terminal elements,
  markers for the left and right structural delimiters (LSD and RSD,
  respectively) and possibly other markup irrelevant to the content of
  the text or its structural analysis.  Occurrences of structural
  delimiters are taken to be properly nested.  We assume only that the
  terminal elements of some corpus can be determined, and not that the
  definition of terminal element correspond to some notion of, say,
  word.  A consequence of this is that markers in a corpus for empty
  elements may be retained, and operated on, even if such markers are
  additional to the original text, and represent part of a hypothesis
  as to the text's linguistic organization.

The following sequences can then be computed from each corpus:
  \begin{flushleft}
    \begin{tabular}[t]{ll}
      $W^{\{l,r\}}$ & the terminal elements \\
      $S^{\{l,r\}}$ & the terminal elements \\ & \hfill and structural delimiters
    \end{tabular}
  \end{flushleft}
So $S$ is the corpus retaining structural annotation, and $W$ is a 
``text only'' version of the corpus.  As each of these is a sequence,
we can pick out elements of each by an index, that is $W^l_n$ will
pick out the $n$th terminal element of the left corpus.  

The following definitions allow us to refer to structural units
(subtrees) within the two corpora.  (We omit the superscript
indicating which corpus we are dealing with.) 

\paragraph{Numbering subtrees}

We number the subtrees in each corpus as follows.  If $S_i$ is the $i$th
occurrence of LSD in $S$ and $S_j$ is the matching RSD of $S_i$, then 
the extent of subtree ($i$) of $S$ is the sequence $S_i \ldots
S_j$. The terminal yield of a subtree is then its extent less any
occurrences of LSD and RSD.  This can be conveniently represented as
the stretch of terminal elements included within a pair of structural
delimiters, i.e.  
\begin{displaymath}
  \mbox{yield}(t) = \langle k, l \rangle 
\end{displaymath}
where $W_k$ is the first element in the extent of $t$ and $W_l$ the
last. We'll refer to a subtree's number as its index. 
Let Subtrees($C$) be the set of yields in $C$. 

\paragraph{Two corollaries}

The following result will be useful later on: for two subtrees from a
corpus, if $t < t'$ then either $t'$ is a subtree of $t$ or there is
no dominance relation between $t$ and $t'$.   

Likewise, we claim that, if a subtree is greater than unary branching,
then it is uniquely identified by its yield.  To see this, suppose
that there are two distinct  subtrees, $t, t'$ such that 
yield($t$) = yield($t'$) or $= \langle i, j\rangle$.  Then, no
terminal element intervenes between $W_i$ and $t$'s LSD, or between
$W_j$ and $t$'s RSD, and the same condition holds of $t'$.  It must
therefore follow that $t$ is a subtree of $t'$ or \textit{vice versa}
and that they are connected by a series of only unary branching
trees.  

\paragraph{Alignment of terminal elements}

We want to compute the minimal set of differences between $W^l$ and
$W^r$, i.e.\ a monotone, bijective partial function $\delta$ defined as
follows:\footnote{Of course, in the general case, such a function may
  not be unique.  It seems a reasonable assumption in the case of
  substantial texts in a natural language that the function will be
  unique (although perhaps empty).} 
\begin{flushleft}
Let $\delta$ be the largest subset of $i \times j$ for  $0 < i \leq
\mbox{length}(W^l)$ and $0 < j \leq \mbox{length}(W^r)$
such that $\delta$ is monotone and bijective, and
\begin{displaymath}
  \begin{array}[t]{lrl}
  \delta(i) = j & \mbox{ if either } &  W^l_i = W^r_j \\
& \mbox{ or }   & 1 < i < \mbox{length}(W^l), \\ && W^l_{i-1} =
W^r_{j-1},\\ &&     1 < j  < \mbox{length}(W^r),  \\ && \mbox{ and } W^l_{i+1} = W^r_{j+1} 

  \end{array}
\end{displaymath}
\end{flushleft}
In other words, $\delta$ records exact matches between the left and
right corpora, or  mismatches involving only a single element, with
exact matches to either side.   This allows minor editorial
differences and choice of markup for terminal elements to have no
effect in overall alignment.  

\paragraph{Aligned subtrees}

We now offer the following definition.  Two trees in $C^l$ and $C^r$
are aligned, if they share the same yield (under the image of
$\delta$), i.e.:

\begin{displaymath}
  \begin{array}[t]{l}
  \langle W^r_{i}, W^r_j  \rangle \in \mbox{Subtrees}(C^r)  \mbox{ and } \\
  \langle W^l_{\delta(i)}, W^l_{\delta(j)}  \rangle  \in \mbox{Subtrees}(C^l)
  \end{array}
\end{displaymath}

Two subtrees are \emph{strictly aligned} if the above conditions hold
and neither tree is a unary branch.  (This definition will be extended
shortly.)  We saw above that, if a tree is not unary branching then its
yield is unique.  

\paragraph{Unary branching}

In the case of unary branching, the inverse of yield will not be a
function. In other words, two subtrees have the same yield.  The
situation is straightforward if both corpora share the same number of
unary trees for some yield: we can pair off subtrees in increasing
order of index.  (Recall that, under dominance, a higher subtree index
indicates domination by a lower index.)  In this case we will say that
the unary trees in question are also strictly aligned.

If the two corpora differ on the number of unary branches relating two
nodes, there is no principled way of pairing off nodes, without
exploiting more detailed, and probably corpus- or markup-specific
information about the contents of the corpora.

\paragraph{Linking to original corpus}

For each of the corpora we assume we can define two functions, one
\textit{terminal location}
will give the location in the original corpus of a terminal element
(e.g.\ a function from terminal indices to, say, byte offsets in a file),
and the other \textit{tree location} will give the location in the
original corpus of a subtree (in terms, say, of byte offsets of the
left and right delimiters).   Tree locations will therefore include
any additional information within the corpus stored between the left
and right delimiters.

\paragraph{Output of the procedure}

The following information may be output from this procedure in the
form of tables
\begin{itemize}
\item of subtree indices indicating strict alignment of two
  trees
\item a table of pairs of sequences of subtree indices indicating
  potential alignment
\item of pairs of terminal element indices, (i.e. the function
  $\delta$) and
\item of single terminal element mismatches, for later processing to
  detect consistent differences in markup.
\item of the results of applying the functions \textit{terminal
    location} and \textit{tree location} to the relevant information
    above.  
\end{itemize}
This output can be thought of as a form of ``stand off'' annotation,
from which other forms of information about the corpora can be
derived.  

\section{A portable implementation}

\begin{figure*}[tb]
  \begin{center}
    \leavevmode
        \psfig{file=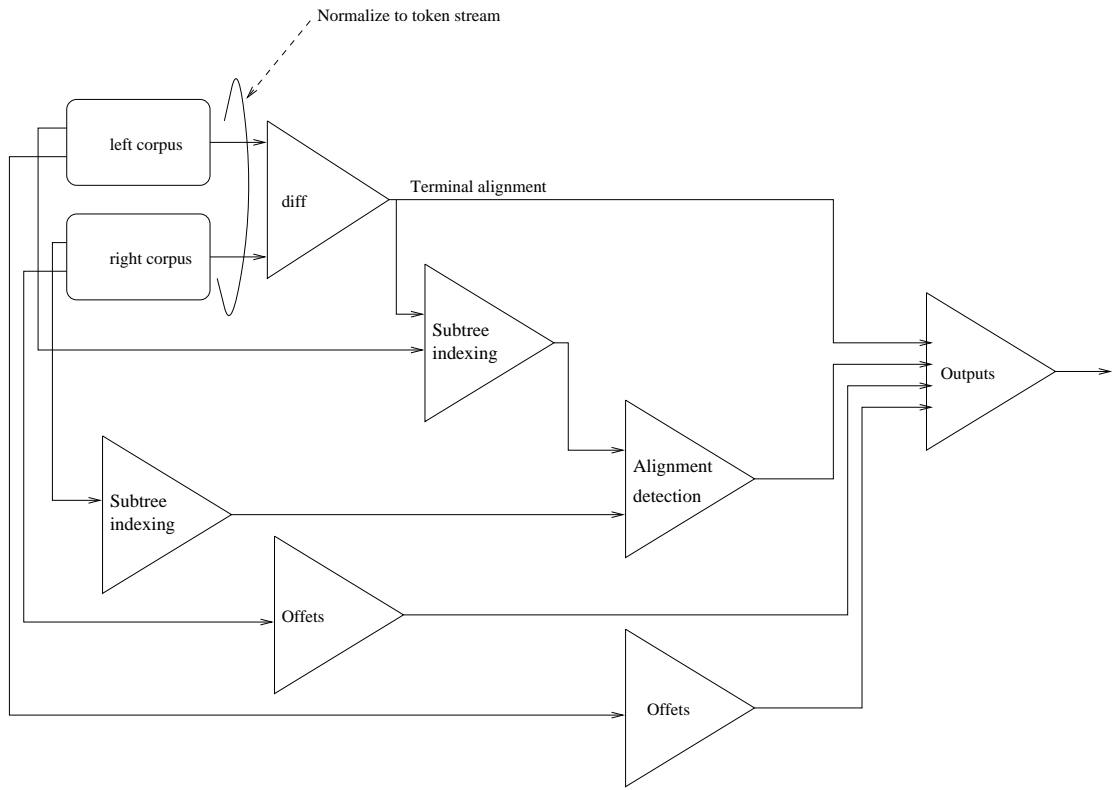} 
    \caption{Overall view of processing}
    \label{fig:arch}
  \end{center}
\end{figure*}
In this section we describe the implementation of the above procedure
which abstracts away from details of the markup used in any particular
corpus.  The overall shape of the implementation is shown in
Figure~\ref{fig:arch}.  The program described here is implemented in
Perl. 

\paragraph{Normalization}  We can abstract away from details of the
markup used in a particular corpus by providing the following
externally defined functions.  

\begin{description}
\item[annotation removal and transformation]
As our procedure works only in terms of terminal elements and
structural annotation, all other information may be removed from a
corpus before processing.  We also take this opportunity to transform
the LSD and RSD used in the corpus into tokens used by the core
processor (that is, \texttt{\{} and \texttt{\}} respectively).  We may
also choose at this point to normalize other aspects of markup known
to consistently differ between the two corpora.  

\item[terminal and tree locations] 
Similarly, separate programs may be invoked to provide tables of byte
offsets of terminals and start- and end-points of trees.  
\end{description}

With these functions in place, we proceed to the description of the
core algorithm.  

\paragraph{Computing minimal differences}

We use the program \texttt{diff} and interpret its output to compute
the function $\delta$.  Specifically we use the Free Software
Foundations \texttt{gdiff} with the options \texttt{--minimal},
\texttt{--ignore-case} and \texttt{--ignore-all-space}, to guarantee
optimal matches of terminals, and allowing editorial decisions that
result in differences in capitalization.  

\paragraph{Subtree indexing and alignment detection}

We use the following for representation of subtrees and the
time-efficient detection of aligned trees.  Trees in the right corpus
(which we can think of as the target) are represented as elements in a
hash table, whose key is computed from the terminal indices of the
start and end of its yield.  Each element in the hash table is a set
of numbers, to allow for the hashing of multiple unary trees to the
same cell in the table.  

In processing the subtrees for the left corpus, we can simply check
whether there is an element in the hash table for the terminal indices
of the yield of the tree in the left corpus under the image of the
function $\delta$.  

\section{An example}
IN this section we give a brief example to illustrate the operations
of the algorithm. The start of the Susanne corpus is shown in the
table here:  

\begin{tabular}[t]{llp{0.2in}ll}
the   & [O[S[Nns:s.  & & \\
Fulton   & [Nns.  & & \\
county   & .Nns]  & & \\
grand   & .  & & \\
jury   & .Nns:s]  & & \\
say   & [Vd.Vd]  & & \\
Friday   & [Nns:t.Nns:t]
\end{tabular}

while the corresponding part of the treebank looks as follows. 
\begin{verbatim}
( (S 
    (NP (DT The) (NNP Fulton) (NNP County) 
                     (NNP Grand) (NNP Jury) )
    (VP (VBD said) 
      (NP (NNP Friday) )
\end{verbatim}

The process of numbering the terminal elements and computing the set
of minimal differences will give rise to a normalized form of the two
corpora something like the following, where the two leftmost columns
come from Susanne, the others from Penn.  (The numbers here have been
altered slightly for the purposes of exposition.) 

\begin{tabular}[t]{llll}
{\it Susanne word} & {\it position} & {\it Penn word} & {\it
  position} \\ 
the       & 2 & the       & 1 \\
Fulton    & 3 & Fulton    & 2  \\
County    & 4 & County    & 3  \\
Grand     & 5 & Grand     & 4  \\
Jury      & 6 & Jury      & 5  \\
\end{tabular}

 Note that the function $\delta$ will in this case map 2 to 1, 3 to 2
and so on.  Note that the whole of this sequence of words is bracketed
off in both corpora.  Accordingly, we will record the existence of a
tree spanning 1 to 5 in the treebank.  The alignment of the
corresponding tree from Susanne will be detected by the noting that
$\delta(2) = 1$ and $\delta(6) = 5$.  

\section{Results of processing on two corpora}

We have processed the entire Susanne corpus and the corresponding
parts of the Penn treebank, and produced tables of alignments for each
pair of marked-up texts. Inputs for this process were a Susanne
file and the corresponding  ``combined'' file from the treebank
(i.e. including part-of-speech information).  
Recalling that the treebank marks up the relationship between
pre-terminal and terminal as a unary tree (and that Susanne doesn't
do this), the treebank regularly contains more trees than Susanne.  

First, a definition: a tree is maximal if it is not part of another
tree within a corpus. 
We ignore maximal trees of depth one in both corpora (as these correspond
to indications of textual units rather than sentence-internal
structural markup).  Each maximal tree containing a tree of greater
than depth one in the treebank may also contain sentence punctuation
which is treated within the structural markup.  As such markup is
typically treated as external to structural annotations within
Susanne, trees containing a sentence and sentence punctuation cannot
be a possible target for alignment across the two corpora. We can take
the number of maximal trees of depth more than one within Susanne as
an indication of the number of trees within the treebank which are
unalignable as a consequence of decisions about markup.  This figure
comes to 2431.

With those considerations,  we report  the following findings:

\begin{itemize}
\item There are 156584 terminal elements in Susanne and of those we find a
total of 145583 (93\%) for which a corresponding element is identified
in the treebank.  The corresponding figure for the treebank is 86\% (of
169782 terminal elements in the treebank).

\item There are 110484 trees in Susanne (including 1952 maximal trees
  of depth one) and so a total of 108532 potentially aligned trees. Of
  these 76011 (70\%) are aligned with trees in the treebank.  

\item There are 301086 trees in the treebank, of which we can
 eliminate 169782 as trees indicating preterminals (which includes
 122174 containing just a textual delimiter), and an estimated further
 2431 as representing trees including sentence punctuation.  This
 gives a total of 128873 (= 59\%) of trees in the treebank possibly
 aligned with those in Susanne are in fact aligned.
\end{itemize}
  
The figures above bear out the impression that trees in the Penn
treebank are more highly articulated than those in Susanne, even
leaving aside the additional structure induced by the treatment of
punctuation and preterminals in the treebank.  

The entire process of computing the above output completes in
approximately fifty minutes on an unloaded Sun SparcStation 20.

\section{Conclusions and Limitations}

We have seen above a formal characterization  and implementation of an
algorithm for determining the extent of agreement between two
corpora.  The core algorithm itself and output formats are completely
independent of the markup used for the different corpora.  The
alignments computed for the Susanne corpus and corresponding portion
of the Penn treebank have been presented and discussed.  

Having computed the alignment of trees across corpora, one option is
to compute (either explicitly or in some form of stand-off annotation)
a corpus combining the information from both sources, thereby allowing
the use of the distinctions made by each corpus at once.  

There are many future experiments of obvious interest, particularly
those to do with examining potential factors in cases of agreement or
disagreement: 

\begin{itemize}
\item analysis of consistency of annotation by markup label

Certain phrase types may be more consistently annotated than others,
so that we can be more confident in our analyses
of such phrases.

\item analysis of consistency of annotation by depth in tree 

From the above discussion we can see that alignment of maximal trees 
approximates 100\%, while that for terminals approximates 90\%.
Therefore (and unsurprisingly) the bulk of disagreement lies somewhere
in between.  Is that disagreement evenly distributed or are there
factors to do with the complexity of analysis at play? 
\end{itemize}

These proposals have to do essentially with formal aspects of markup.
Other, perhaps more interesting questions, touch on the linguistic
content of analyses, and whether for example particular linguistic
phenomena are associated with divergence between the corpora. 

The assumption that trees within corpora are strictly nested 
represents an obvious limitation on the scope of the algorithm.  In
cases where markup is more complex, other strategies will have to be
developed for detecting agreement between corpora.  That said, the
class of markup for which the algorithm presented here is applicable
is very large, including perhaps most importantly normalized forms of
SGML \cite{goldfarb}, for example that proposed by
\cite{thompson+mckelvie}.  

\section{Acknowledgements}

I would like to thank Chris Brew for
  conversations on the topics discussed in this paper and the
  anonymous referees for their comments.  This work was
  funded by the UK DTI/SERC-funded SALT Programme as part of the
  project \textit{ILD: The Integrated Language Database}.

\end{document}